\documentclass[]{bytedance_seed}
\usepackage{helvet}

\usepackage{amsmath} 
\usepackage{natbib}
\usepackage{graphicx}
\usepackage{subcaption}

\usepackage[toc,page,header]{appendix}
\usepackage[utf8]{inputenc} 
\usepackage[T1]{fontenc}    
\usepackage{hyperref}       
\usepackage{url}            
\usepackage{booktabs}       
\usepackage{lmodern}        
\usepackage{amsfonts}       
\usepackage{nicefrac}       
\usepackage{microtype}      
\usepackage{wrapfig} 

\usepackage{amssymb}  
\usepackage{fontawesome}  
\usepackage{url}  

\usepackage{minitoc}

\usepackage{booktabs}
\usepackage{array}
\usepackage{etoolbox}

\definecolor{lightblue}{RGB}{200, 230, 255}  
\definecolor{headerblue}{RGB}{150, 200, 255} 
\definecolor{tableblue}{RGB}{238, 249, 253}

\title{LLM-I: LLMs are Naturally Interleaved \\Multimodal Creators}
\author[1,2]{Zirun Guo}
\author[2]{Feng Zhang}
\author[2]{Kai Jia}
\author[1]{Tao Jin}

\affiliation[1]{Zhejiang University}
\affiliation[2]{ByteDance, BandAI}

\abstract{
\begin{abstract}

We propose LLM-Interleaved (\textbf{LLM-I}), a flexible and dynamic framework that reframes interleaved image-text generation as a tool-use problem. LLM-I is designed to overcome the ``one-tool'' bottleneck of current unified models, which are limited to synthetic imagery and struggle with tasks requiring factual grounding or programmatic precision. Our framework empowers a central LLM or MLLM agent to intelligently orchestrate a diverse toolkit of specialized visual tools, including online image search, diffusion-based generation, code execution, and image editing. The agent is trained to select and apply these tools proficiently via a Reinforcement Learning (RL) framework that features a hybrid reward system combining rule-based logic with judgments from LLM and MLLM evaluators. Trained on a diverse new dataset using four different model backbones, LLM-I demonstrates state-of-the-art performance, outperforming existing methods by a large margin across four benchmarks. We also introduce a novel test-time scaling strategy that provides further performance gains.
\end{abstract}
}

\date{\today}
\correspondence{Zirun Guo at \email{zrguo.cs@gmail.com}}

\checkdata[Project Page]{\url{https://github.com/ByteDance-BandAI/LLM-I}}

\begin{document}
\maketitle


\section{Introduction}

AI is shifting from single-modality systems to multimodal ones that can process mixed data like text, images, and sound. A key frontier is interleaved image-text generation~\citep{ge2024seed, tian2024mminterleavedinterleavedimagetextgenerative, xie2025show, zhou2025opening, xia2025mmie,chen2025interleaved}: producing a coherent, alternating sequence of text and images from a single prompt. However, the task is technically demanding, requiring high-fidelity text and images with strict cross-modal consistency. This involves maintaining narrative coherence, consistent visual style and entities, and strong semantic alignment between each image and its accompanying text.

To address these challenges, the research community has largely converged on two dominant architectural paradigms for interleaved image-text generation. The first, a two-stage or compositional approach, leverages the distinct strengths of separate, state-of-the-art models~\citep{zhou2025opening} or add decoders~\citep{ge2024seed} after the text generation. In this paradigm, a powerful LLM, such as GPT-4o~\citep{hurst2024gpt}, acts as a high-level reasoning engine. It interprets the user's request to produce a sequence of textual narratives, which are then passed to a separate, high-fidelity text-to-image diffusion model, such as DALL-E~\citep{betker2023improving} or Seedream~\citep{gao2025seedream}, for visual synthesis. However, it often suffers from a ``semantic gap'', where the LLM's textual representation of a desired image may not perfectly align with the diffusion model's interpretation, leading to inconsistencies. Furthermore, these systems lack flexibility, as they are typically restricted to generating a fixed number of images per response.

Seeking to close this gap and achieve greater architectural elegance, a significant research effort has been directed towards developing unified, end-to-end models~\citep{xie2025show, zhou2025transfusion, deng2025emerging} that handle both multimodal understanding and generation within a single, integrated framework. Despite their notable advancements, current unified models for interleaved generation suffer from a critical and largely unaddressed limitation: the ``one-tool'' bottleneck. While these unified models excel at generating novel, high-fidelity synthetic imagery from textual prompts, they are inherently ill-suited for tasks that require factual grounding such as real-world images or programmatic precision such as data analysis and visualizations. This architectural commitment creates a rigid system that forces a single tool to solve all visual generation problems, regardless of its suitability. This ``one-tool'' bottleneck reflects a deeper paradigm choice in AI development: the pursuit of an ``omniscient solver'' that embeds all knowledge within its parameters, rather than a ``proficient tool-user'' that knows how to leverage external resources. The latter approach is inherently more flexible, scalable, and robust. A tool-augmented system can be easily updated with new capabilities by simply adding a new tool to its repertoire, whereas a monolithic model requires complete and computationally prohibitive retraining to acquire new skills.

In this paper, we introduce LLM-Interleaved (\textbf{LLM-I}), a flexible and dynamic framework that employs an LLM or MLLM as an agentic planner. This central agent leverages its sophisticated reasoning and multimodal understanding capabilities to intelligently orchestrate a diverse suite of external, specialized visual tools for image generations. Our framework equips the central agent with a toolkit of four distinct and complementary visual tools which are \textit{online image search, diffusion-based generation, code generation and execution}, and \textit{image edit tool}. To ensure the agent uses these tools proficiently, we develope a Reinforcement Learning (RL) framework that incorporates a hybrid reward design, combining rule-based rewards and LLM and MLLM judges. We build a diverse dataset for training and evaluate LLM-I using four different backbone models, finding that it outperforms state-of-the-art methods by a large margin across four benchmarks. Additionally, we propose a novel test-time scaling strategy that improves performance even further.

We summarize our key contributions as follows:
\begin{enumerate}
\item \textbf{Novel Framework for Interleaved Generation}: We propose a new and flexible paradigm, LLM-I, for interleaved image-text generation. Our framework recasts the LLM/MLLM not as an end-to-end generator but as an intelligent agent that orchestrates a toolkit of external, specialized visual models. This approach decouples high-level reasoning from low-level synthesis, enabling unprecedented flexibility and context-appropriateness in the generated multimodal content.

\item \textbf{New Dataset and Benchmark}: We introduce a diverse dataset and difficult benchmark for interleaved image-text generation. Our work moves beyond the scope of previous datasets by requiring multiple forms of images, including retrieved real-world photos, synthetic visuals, and programmatic visualizations.

\item \textbf{Strong Performance}: LLM-I outperforms previous SOTA methods by a large margin across four benchmarks. Thourgh test-time scaling, the performance is further improved.
\end{enumerate}

\section{Related Work}

\textbf{Interleaved Image-Text Generation. }
While current MLLMs, such as the QwenVL~\citep{bai2025qwen2} and InternVL~\citep{zhu2025internvl3} series, excel at processing interleaved image-text inputs, they lack the capability for interleaved generation. Two primary approaches have emerged to address this limitation. The first involves leveraging an external image decoder or diffusion model, as seen in models like NExT-GPT~\citep{wu2024next} and SEED-X~\citep{ge2024seed}. These methods typically optimize a set of learnable visual tokens that serve as input for a diffusion-based image decoder or directly input all the texts into the diffusion model. The second category consists of unified multimodal models that either integrate an autoregressive model with a diffusion model~\citep{zhou2025transfusion, xie2025show} or are entirely autoregressive~\citep{team2024chameleon, chern2024anole} to achieve unified training and alignment. However, a significant drawback of both paradigms is their inherent unsuitability for tasks requiring factual grounding, such as generating photorealistic images of specific entities, or programmatic precision, such as data analysis and visualization. Diverging from these methods, our approach reframes the LLM or MLLM as an agentic planner that orchestrates four external tools. This tool-augmented framework allows for the creation of a wide range of visual content, from photorealistic and creative imagery to accurate data visualizations, thereby overcoming the key weaknesses of prior generative systems.

\textbf{Reinforcement Learning. }
RL has become a crucial component in developing the latest generation of large models~\citep{guo2025deepseek}, often yielding superior generalization capabilities compared to purely supervised methods. While Proximal Policy Optimization (PPO)~\citep{schulman2017proximalpolicyoptimizationalgorithms} is the most common algorithm for fine-tuning LLMs, its reliance on a value model has spurred the popularity of value-free alternatives like GRPO~\citep{shao2024deepseekmathpushinglimitsmathematical} and DAPO~\citep{yu2025dapoopensourcellmreinforcement}. Although many recent works have successfully applied these algorithms to enhance the reasoning abilities of LLMs and MLLMs~\citep{zheng2025group, guo2025observe, hong2025apo}, our research explores a different direction. Instead of focusing on reasoning, we investigate how RL can be used to improve multimodal alignment, the ability to intelligently use tools, and the overall quality of generated reports.

\textbf{Tool Usage of LLMs. }
The ability of LLMs to utilize external tools~\citep{feng2025retool, wu2025mmsearch} has significantly expanded their capabilities, transforming them from simple text generators into sophisticated agents capable of reasoning, decision-making, and task automation across various domains. For instance, proprietary models like the OpenAI o3~\citep{o3} and DeepResearch~\citep{deepresearch} model can leverage various tools for web search, code execution, and image processing. Similarly, Gemini 2.5 Pro~\citep{comanici2025gemini} and its DeepResearch~\citep{geminideepresearch} can call external tools for functions like code execution, web search, or file processing. In the open-source community, projects such as Search-o1~\citep{li2025search} and Openthinkimg~\citep{su2025openthinkimg} have also demonstrated the impressive performance improvements of tool-augmented LLMs and MLLMs. Building on these advancements, RL training can further enhance this capability, enabling an LLM to intelligently select the appropriate tool to use, making it possible to address a wider and more complex range of problems.
\section{Methodology}
\label{sec:method}

\subsection{Tool Usage}
\subsubsection{Motivation}
As we discussed above, current methods~\citep{chern2024anole, wu2024next, zhou2025transfusion, xie2025show} are locked into a single mode of creation, limiting the scope, factuality, and utility of the narratives they can produce. It is instructive to consider how humans approach a similar task, such as authoring a blog post or a technical report. When a writer needs to insert an image, they rarely create it from scratch. Instead, they act as an intelligent agent, selecting the best external tool for the job. If they need a picture of the Eiffel Tower, they use a search engine to find a real photograph. To display quarterly sales data, they would use software like PowerPoint or a coding library to generate a precise chart. They might also use an image editing tool like Photoshop to make adjustments, such as cropping a photo, adjusting its colors, or adding annotations to highlight key information. This human workflow is not monolithic; it is dynamic, flexible, and tool-centric. The writer's primary skill is not drawing but reasoning and orchestrating a diverse set of specialized tools to achieve their goal.

Therefore, we argue that a paradigm that mimics this human-like, tool-using strategy holds significant advantages over current monolithic models. An AI system that can intelligently invoke external tools is inherently more flexible, scalable, and robust. It can ground its generations in factual reality by searching the web, provide precise data visualizations through code execution, and still retain the ability for other tasks. This approach directly overcomes the ``one-tool'' bottleneck, moving beyond the limited ``omniscient solver'' paradigm towards a more powerful and practical ``proficient tool-user''.

\subsubsection{Toolkit}
Motivated by this insight, we introduce a flexible and dynamic framework where an LLM or MLLM serves as an agentic planner. We empower this central agent to intelligently orchestrate a suite of distinct visual tools to construct rich, interleaved content. Specifically, our framework equips the agent with capabilities for online image search, diffusion-based generation, code execution for data visualization, and image editing. 
\begin{enumerate}
    \item \textbf{Online Image Search}: Invoked for requests demanding factual grounding, such as specific real-world entities, landmarks, or current events. This tool ensures visual authenticity and provides access to up-to-date information beyond the model's training data cutoff. In our paper, we use Google Search API~\citep{serpapi}.
    \item \textbf{Diffusion-based Generation}: Selected for tasks requiring the creative synthesis of novel or abstract concepts, or complex compositions that do not exist in reality. We support Seedream 3.0~\citep{gao2025seedream} in our paper.
    \item \textbf{Code Execution}: Utilized primarily for generating data visualizations like charts, graphs, and plots from structured data. We use Python as the programming language and build a controlled sandbox environment.
    \item \textbf{Image Editing}: Engaged to perform modifications on existing visual content, whether inputted,  retrieved or generated. We support Seededit 3.0~\citep{wang2025seededit} in our project.
\end{enumerate}

\subsubsection{How to call a tool?}
To empower the LLM to dynamically orchestrate our suite of visual tools, we design a robust and flexible tool invocation framework. Instead of complex, multi-turn interactions or fine-tuning on specific API call formats, our approach is guided by a system prompt that instructs the model to embed a specific placeholder tag wherever a visual element is required in the narrative. This method allows the LLM to autonomously decide when and how to use a tool within a single generative pass.

The core of our framework is the structured tag, \texttt{<imgen>\{...\}</imgen>}, which encapsulates all the necessary information for generating or retrieving an image. When the LLM determines that an image is needed, it generates this tag in the following JSON-like format: 

\texttt{<imgen>\{"source":"<source type>", "description":"<general title>", "params":\{...\}\}</imgen>}

For search, the params contains a single key \textit{query} which holds a practical and concise search string for a web image search engine. For diffusion, it contains the key \textit{prompt}, which provides a descriptive text prompt for the generative model. For code, it contains the key \textit{code} which holds the raw Python code snippet required to generate a plot or visualization. For edit, it contains two keys, \textit{img index}, the 0-based index of a previously image in the sequence to be modified, and \textit{prompt}, a textual instruction describing the desired edit.

When the tag is detected in the generated sequence, a parser processes this output, identifies each tag, and dispatches a call to the corresponding external tool using the provided parameters. The tag is then replaced in the text with the image returned by the tool, resulting in the final, seamless multimodal document.

\begin{figure}
    \centering
    \includegraphics[width=0.9\linewidth]{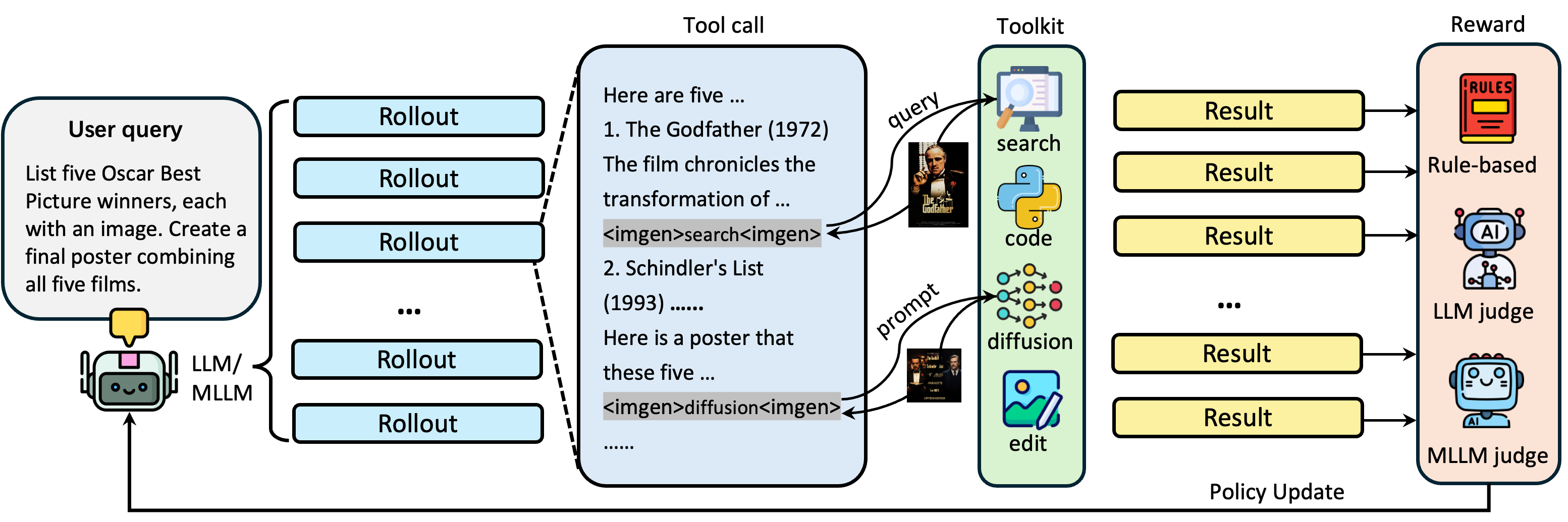}
    \caption{
Overview of the LLM-I framework.
}

    \label{fig:main}
\end{figure}

\subsection{RL Recipe}
\subsubsection{Dataset Construction}\label{sec:data}
To effectively train our model to master the agentic tool-use framework, we first construct a high-quality RL dataset. The central design philosophy is ``tool-oriented'', aimed at teaching the model to invoke a diverse set of tools under various constraints. The dataset is bifurcated into two primary categories: text-only inputs and text-and-image inputs.

The generation process is automated using Gemini 2.5 Pro~\citep{comanici2025gemini}. We guide prompt creation through a categorical scaffolding system that defines the target tool(s), a pre-designed specific theme for the tool, an image count which implicitly specifies how many images should be given in the response, and a difficulty level (low, medium, high). A crucial principle is that all generated prompts are implicit; they describe a desired outcome and image number that necessitates a specific tool without ever naming it, thereby encouraging the model to reason about tool selection and image number. To counteract the agent's potential aversion to more error-prone tools during RL (a form of reward hacking), we deliberately increase the representation of prompts requiring code and search, which have higher failure rates than the more predictable diffusion tool.

For the text-and-image input subset, the generation process is adapted to produce both an instructional prompt and a  textual description of required input images. This description is then used to synthesize the image via Nano Banana~\citep{nanobanana}. The composition of this subset is slightly weighted towards the edit tool, as its function is inherently tied to modifying existing visual content.

To ensure the quality and fidelity of the entire dataset, we implement a rigorous multi-stage validation pipeline using GPT-4o~\citep{hurst2024gpt} as an independent adjudicator. This pipeline verifies three key aspects for each sample: the consistency of the intended image count, the appropriateness of the designated tool for the given instruction, and, for the text-and-image subset, the cross-modal alignment between the synthesized input image and its textual description. Any sample that fails a validation check is discarded, resulting in a high-quality, unambiguous dataset optimized for robust RL-based agent training. Finally, we get around 4k samples.

A critical feature of this dataset is the annotation of each prompt with an image num constraint. This metadata guides the RL training process by specifying the rules of image generation for each task (Section~\ref{sec:reward}). The constraint falls into one of four categories: images are disallowed (-1), their use is unconstrained (0), a precise quantity $n$ is required ($n$ > 0), or at least one image is mandatory (\texttt{Inf}).

\subsubsection{Reward}\label{sec:reward}
With the instruction dataset in place, we employ an RL strategy to fine-tune the model's ability to appropriately call and parameterize the visual tools. Our approach is distinguished by a multi-faceted reward function that combines deterministic rules $R_{rule}$ with sophisticated judgments from both LLM $R_{llm}$ and MLLM $R_{mllm}$. This composite reward signal not only provides a holistic assessment of the generated output but also decreases reward hacking.

The first component is a deterministic, rule-based reward $R_{rule}$ that enforces adherence to generation constraints and ensures the correctness of the \texttt{<imgen>} tag format. In Section~\ref{sec:data}, we set a required image number $N_{req}$ for each single item. For categorical constraints, the reward is binary. When images are disallowed ($N_{req}=-1$) or when at least one is required ($N_{req}=\inf$), the model receives a score of 1 for compliance and 0 for violation. When there is no constraint ($N_{req}=0$), the score is always 1, as any output is considered valid. For quantitative constraints where a precise number of images $n$ is required ($N_{req}=n$), the reward is designed to penalize both under- and over-generation:

\begin{equation}
    R_{rule} = 
\begin{cases} 
    \frac{N_{gen}}{N_{req}} & \text{if } 0 \le N_{gen} \le N_{req} \\
    \max(0, 1 - \alpha \cdot (N_{gen} - N_{req})) & \text{if } N_{gen} > N_{req}
\end{cases}
\end{equation}
where $N_{gen}$ is the number of generated images in the model response and $\alpha$ is the penalty factor of extra images which is set to 0.3 by default.

The second component $R_{llm}$ leverages an external LLM as a judge to assess the quality of the language and the logic of the tool invocation. This judge evaluates two criteria on a 1-to-5 scale: (i) the fluency, coherence, and relevance of the textual narrative, and (ii) the quality of the tool-use tags, including the naturalness of their placement and the semantic appropriateness of the chosen source and params.

The third reward component $R_{mllm}$ employs an MLLM to evaluate the final interleaved output. After the images are generated and integrated, this judge scores three key aspects of multimodal quality on a 1-to-5 scale: (i) the technical and aesthetic quality of the image itself, (ii) the semantic alignment between the image and its surrounding text, and (iii) the relevance of the image to the overall task objective.

The scores from the LLM and MLLM judges are normalized to a [0, 1] range. The final reward signal $R$, is then composed from all three components. Notably, the rule-based reward $R_{rule}$, acts as a multiplicative gate on the MLLM reward $R_{mllm}$. This formulation ensures that visual quality is considered only if the model has first satisfied the explicit image count constraint. The composite reward is thus defined as:
\begin{equation}
    R = w_{rule}R_{rule} + w_{llm}R_{llm} + w_{mllm}R_{mllm}R_{rule}
\end{equation}
where $w_{rule}$, $w_{llm}$, and $w_{mllm}$ are the trade-offs between the three losses.

\begin{figure}
    \centering
    \includegraphics[width=0.95\linewidth]{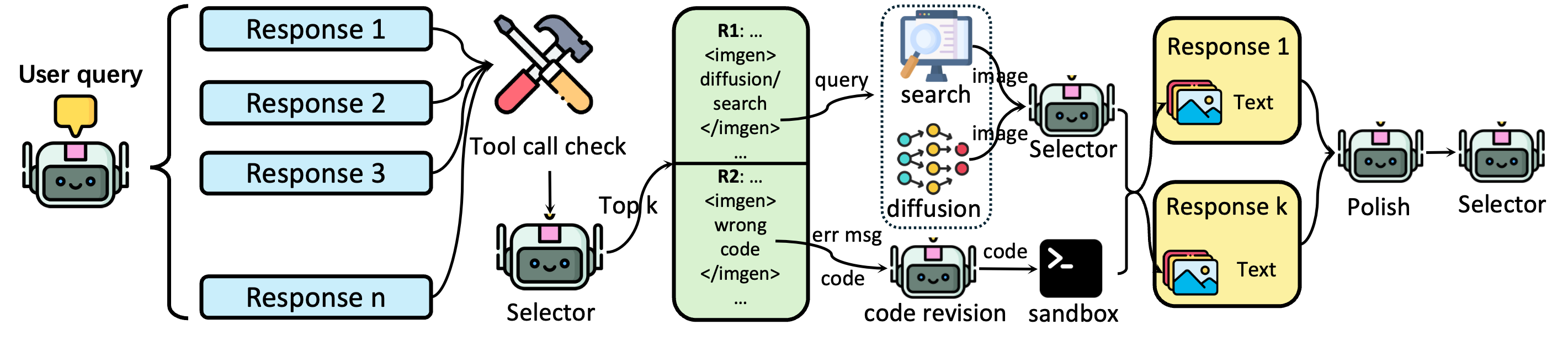}
    \caption{Overview of test-time scaling framework for LLM-I.}

    \label{fig:tts}
\end{figure}

\subsection{Test-time Scaling}\label{sec:tts}
To further enhance the performance of our agentic framework, we introduce a \textit{test-time scaling}~\citep{snell2025scaling, muennighoff2025s1simpletesttimescaling} paradigm that leverages additional computational resources during inference. The goal is to improve both the reliability of tool usage and the overall quality of the final multimodal response. The workflow is illustrated in Figure~\ref{fig:tts}.

Given a user query, the model first generates multiple complete candidate responses through stochastic sampling. Each candidate may contain tool calls (\textit{e.g.,} search, diffusion, code, or editing), interleaved with natural language. The initial $n$ candidates are passed through a ``Tool Call Check'' filter. This stage validates the structural integrity and executability of the tool invocations. Responses with malformed or failed tool calls are discarded. From the pool of successful candidates, a selector model (an LLM/MLLM) evaluates their overall quality and relevance to the prompt, selecting the top-k most promising responses for further enhancement. Then, each of the $k$ selected candidates undergoes a targeted enhancement process based on the type of tool used. When a response requests an image, we concurrently query both the online image search module and the diffusion model. The resulting candidates are evaluated by an MLLM, which selects the most semantically aligned option. If code execution fails, the erroneous code and the associated error message are provided to a model. The model revises the code, which is then re-executed in a sandboxed environment until a valid visualization is obtained or exceeding a fixed number of attempts. After the enhancement, the $k$ refined interleaved multimodal responses are passed to an MLLM for polishing. This step improves the coherence and alignment between modalities, ensuring that visual outputs are seamlessly integrated with textual explanations. Finally, a selector model ranks the polished candidates and chooses the single best response as the final output.

\section{Benchmark}
To rigorously assess a model's capability in generating sophisticated, interleaved text-image reports, we develop a new benchmark. This is motivated by two primary limitations we observe in existing public benchmarks~\citep{liu2024holistic, zhou2025opening, chen2025interleaved}.

First, current benchmarks often feature overly simplistic and generic prompts, such as ``Generate a travel guide to Beijing with text and images.'' The tasks in such benchmarks do not necessitate deep reasoning, and the requested images are often decorative rather than integral to the content (shown in Figure~\ref{fig:isgexp1}). These images typically have low informational density, are stylistically uniform (\textit{e.g.,} lifestyle photos), and can be adequately produced by standard diffusion models without complex planning. Consequently, they fail to test a model's ability to generate meaningful, context-aware visuals that are essential for a high-quality report.

Second, the evaluation protocols of existing benchmarks rely heavily on subjective metrics. They commonly employ models like GPT-4o to score outputs based on broad criteria such as ``text-image alignment,'' ``text quality,'' and ``image quality.'' This approach is problematic, as LLMs tend to assign forgivingly high scores even to suboptimal outputs. In our preliminary tests, we observe instances where a model fails to generate an image and instead provides only a textual description, yet still receives a favorable score from the GPT-4o evaluator. This highlights the unreliability of using vague, subjective rubrics for evaluation.

To overcome these challenges, our benchmark introduces a new paradigm for both task design and evaluation. We reframe the task of interleaved generation as a ``mini-project''. Each prompt in our benchmark provides background context or specific data. The tasks are designed to demand images with high informational value and stylistic diversity, moving beyond simple photographic illustrations. The required images include visuals like data analysis, scientific illustrations, and creative content. In this framework, images are not merely supplementary; they are an indispensable component of the report, carrying critical information that is synergistic with the text. The goal is to ensure that each image serves a distinct purpose, reflecting a genuine user need for visual information. We present four samples in Figure~\ref{fig:llmexp1}, \ref{fig:mllmexp1}, \ref{fig:mllmexp2}, and \ref{fig:llmexp2}.

To address the issue of subjective evaluation, we transition from broad rubrics to a sample-specific, objective evaluation protocol. Instead of asking an LLM for a holistic quality score, we design a unique set of concrete and verifiable criteria for each ``mini-project" sample. For instance, for a report on sales trends, the evaluation criteria include specific, verifiable checks such as ``Does the report accurately generate a line chart for sales from 2014 to 2025 with correct points and labels according to the provided data?'' For each sample in our benchmark, we define 10 distinct evaluation metrics. We utilize GPT-4o to assess the generated report against these specific rules, assigning a score on a three-point scale: 0 (requirement not met), 1 (partially met), or 2 (fully met). This method transforms the evaluation from a subjective assessment into a more objective and reliable measurement of a model's capabilities.

Our final benchmark is concise yet comprehensive, comprising 30 meticulously designed and manually vetted samples. These samples cover a diverse range of topics and user requirements, with 18 being text-only inputs and 12 being multi-modal inputs. We deliberately emphasize ``quality over quantity''. The compact size of 30 samples is a strategic choice to facilitate rigorous and manageable human evaluation. Our approach ensures that each sample can be carefully analyzed, enabling a deeper and more accurate understanding of model performance.
\begin{table}[]
\caption{Results on the OpenING benchmark. $\dagger$ is evaluated with text-only input samples. IT Coherency refers to image-text conherency and MS consistency means multi-step consistency. Qwen series are all evaluated with tools.}
\label{tab:opening}
\resizebox{\columnwidth}{!}{%
\begin{tabular}{@{}l|ccccccc|c@{}}
\toprule
Model & Completeness & Quality & Richness & Correctness & Human Alignment & IT Coherency & MS Consistency & Overall \\ \midrule
GPT4o+DALLE3 & 8.66 & 8.01 & 7.42 & 7.98 & 8.77 & 8.15 & 8.38 & 8.20 \\
Gemini+FLUX & 7.58 & 7.26 & 6.48 & 7.03 & 7.98 & 6.98 & 7.33 & 7.23 \\\midrule
NExT-GPT & 3.89 & 4.25 & 3.35 & 3.61 & 5.35 & 3.32 & 3.85 & 3.95 \\
Show-o & 4.37 & 4.79 & 3.83 & 3.76 & 5.78 & 4.04 & 4.33 & 4.41 \\
SEED-X & 5.65 & 6.07 & 4.92 & 5.77 & 7.03 & 5.72 & 5.72 & 5.84 \\
Anole & 6.27 & 6.02 & 5.28 & 5.06 & 6.91 & 4.90 & 5.81 & 5.75 \\\midrule
Qwen2.5-VL-7B & 2.97 & 3.90 & 2.50 & 3.07 & 4.37 & 2.03 & 3.82 & 3.24 \\
Qwen2.5-VL-32B & 6.78 & 6.82 & 5.89 & 6.34 & 7.25 & 5.69 & 7.15 & 6.56 \\
\rowcolor{tableblue} MLLM-I-7B & 6.00 & 6.75 & 5.53 & 5.85 & 7.24 & 5.85 & 6.50 & 6.25 \\
\rowcolor{tableblue} MLLM-I-32B & 8.35 & 8.07 & 7.48 & 7.79 & 8.44 & 7.35 & 8.38 & 7.98 \\\midrule
Qwen3-4B-Instruct$\dagger$ & 6.26 & 6.88 & 5.55 & 6.09 & 6.95 & 5.11 & 6.86 & 6.24 \\
Qwen3-30B-Instruct$\dagger$ & 8.05 & 7.63 & 7.09 & 7.56 & 8.12 & 6.90 & 8.13 & 7.64 \\
\rowcolor{tableblue}LLM-I-4B$\dagger$ & 8.63 & 8.03 & 7.54 & 8.03 & 8.69 & 7.87 & 8.45 & 8.18 \\
\rowcolor{tableblue}LLM-I-30B$\dagger$ & 9.19 & 8.44 & 8.08 & 8.61 & 8.99 & 8.40 & 8.91 & 8.66 \\ \bottomrule
\end{tabular}%
}
\end{table}

\begin{table}[h!]
\caption{Results on the LLMI-Bench. $\dagger$ is evaluated with text-only input samples. Tool Acc refers the success rate of tool invocation.}
\label{tab:llmibench}
\centering
\resizebox{0.5\columnwidth}{!}{%
\begin{tabular}{@{}l|ccc|c@{}}
\toprule
Model                    & Rubric & Human & Tool Acc & Overall \\ \midrule
GPT-5 wTool              & 53.8   & 48.3  & 28.1     &  43.4       \\
GPT-4o wTool             & 70.4   & 62.8  & 67.9     &  67.0       \\
Anole                    & 27.4   & 18.2  & -        &  22.8       \\\midrule
Qwen2.5-VL-7B wTool      & 28.5   & 19.3  & 44.3     &  30.7       \\
Qwen2.5-VL-32B wTool     & 58.9   & 51.1  & 93.4     &  67.8       \\
Qwen2.5-VL-72B wTool     & 73.1   & 59.8  & 60.1     &  64.3       \\
\rowcolor{tableblue}MLLM-I-7B                & 67.1   & 61.9  & 97.4     &  75.5       \\
\rowcolor{tableblue}MLLM-I-32B               & 92.5   & 82.1  & 99.2     &  91.3       \\\midrule
Qwen3-4B-Instruct wTool$\dagger$  & 73.6   & 62.3  & 68.7     &  68.2       \\
Qwen3-30B-Instruct wTool$\dagger$ & 81.4   & 69.2  & 83.1     &  77.9       \\
\rowcolor{tableblue}LLM-I-4B$\dagger$                & 88.9   & 72.9  & 100.0    &  82.3       \\
\rowcolor{tableblue}LLM-I-30B$\dagger$                & 94.8   & 83.3  & 100.0    &  92.7       \\ \bottomrule
\end{tabular}}
\end{table}

\section{Experiments}

\subsection{Setup}

\textbf{Data and Benchmarks:}
We train our model using the data constructed in Section~\ref{sec:data}, which is split into a training set and an in-domain test set containing over 200 samples. For a comprehensive evaluation, we utilize this in-domain test set along with three out-of-domain (OOD) benchmarks. On the in-domain set, we employ the same metrics used during training: a rule-based metric, LLM-based judgments, and MLLM-based judgments. For OOD evaluation, we use the public OpenING benchmark~\citep{zhou2025opening}, which has over 2,000 samples, and adopt the seven metrics from the original paper. Besides, we use the public benchmark ISG~\citep{chen2025interleaved}, which has over 1,000 samples, and adopt the four metrics from the original paper. Additionally, we introduce our novel and much more difficult LLMI-Bench, whose manageable size enables a multifaceted evaluation through a rubric-based scoring rate from GPT-4o, a rule-based metric to measure tool-call success, and a rigorous human evaluation. For this human assessment, we design a five-point Likert scale with detailed criteria for each point and calculate the final metric as the average overall scoring rate.

\begin{table}[]
\centering
\caption{Results on the ISG benchmark. $\dagger$ is evaluated with text-only input samples.}
\label{tab:isg}
\resizebox{0.45\textwidth}{!}{%
\begin{tabular}{@{}l|cccc@{}}
\toprule
Model & Structural & Holistic & Block & Image \\ \midrule
Show-o & 0.295 & 2.329 & 1.962 & 0.078 \\
Anole & 0.000 & 2.810 & - & - \\
Gemini+SD3 & 0.385 & 5.827 & 3.081 & 0.113 \\
ISG & 0.871 & 6.262 & 5.515 & 0.574 \\\midrule
Qwen2.5-VL-7B & 0.085 & 4.932 & 1.152 & 0.016 \\
Qwen2.5-VL-32B & 0.221 & 6.354 & 2.105 & 0.088 \\
\rowcolor{tableblue}MLLM-I-7B & 0.607 & 6.381 & 3.584 & 0.274 \\
\rowcolor{tableblue}MLLM-I-32B & 0.776 & 8.112 & 5.722 & 0.419 \\\midrule
Qwen3-4B$\dagger$ & 0.068 & 5.621 & 1.621 & 0.086 \\
Qwen3-30B$\dagger$ & 0.267 & 7.848 & 3.811 & 0.267 \\
\rowcolor{tableblue}LLM-I-4B$\dagger$ & 0.881 & 8.413 & 7.701 & 0.511 \\
\rowcolor{tableblue}LM-I-30B$\dagger$ & 0.971 & 8.492 & 8.291 & 0.618 \\ \bottomrule
\end{tabular}%
}
\end{table}

\begin{table}
\centering
\caption{Results on the test set of the dataset. $\dagger$ is evaluated with text-only input samples. IT means image-text and IQ means image-question. Qwen series are all evaluated with tools.}
\label{tab:testset}
\resizebox{0.95\textwidth}{!}{%
\begin{tabular}{@{}l|c|cc|ccc|c@{}}
\toprule
Model & Image num & Text Quality & Tag Quality & Image Quality & IT Alignment & IQ Alignment & Overall \\ \midrule
Qwen2.5-VL-7B & 13.2 & 3.1 & 1.5 & 3.7 & 2.8 & 2.9 & 25.9 \\
Qwen2.5-VL-32B & 54.3 & 3.8 & 2.0 & 3.8 & 3.6 & 3.5 & 52.7 \\
\rowcolor{tableblue}MLLM-I-7B & 88.7 & 4.0 & 3.4 & 3.9 & 3.9 & 3.7 & 70.6 \\
\rowcolor{tableblue}MLLM-I-32B & 95.1 & 4.7 & 4.3 & 4.1 & 4.2 & 4.3 & 85.2 \\\midrule
Qwen3-4B-Instruct$\dagger$ & 46.5 & 4.5 & 2.9 & 4.0 & 3.9 & 3.9 & 57.7 \\
Qwen3-30B-Instruct$\dagger$ & 55.3 & 4.8 & 4.0 & 3.9 & 3.9 & 3.9 & 68.7 \\
\rowcolor{tableblue}LLM-I-4B$\dagger$ & 88.6 & 4.8 & 4.6 & 4.2 & 4.2 & 4.3 & 85.2 \\
\rowcolor{tableblue}LLM-I-30B$\dagger$ & 93.0 & 4.9 & 4.8 & 4.3 & 4.6 & 4.6 & 89.9 \\ \bottomrule
\end{tabular}%
}
\end{table}

\textbf{Training: }
We conduct experiments using four different backbones, covering both LLMs and MLLMs. They include Qwen3-4B-Instruct, Qwen3-30B-Instruct (MoE model), Qwen2.5-VL-7B, and Qwen2.5-VL-32B. For MoE model, we use GSPO~\citep{zheng2025group} as the RL algorithm while for others we use GRPO~\citep{shao2024deepseekmathpushinglimitsmathematical}. We use a batch size of 32 with a cosine learning rate scheduler where the initial learning is set to 1e-6, minimum learning rate ratio is set to 0.01, and the warm-up step is 5. Following \citet{yu2025dapoopensourcellmreinforcement}, we use the token-level loss. For GSPO, the clipping ratios are set to 3e-4 (low) and 4e-4 (high). For judgement, we use Qwen3-235B-Instruct-2507~\citep{yang2025qwen3} as the LLM judge and Qwen2.5-VL-72B-Instruct~\citep{bai2025qwen2} as the MLLM judge. The reward trade-off coefficients are set to $w_{rule}=0.2$, $w_{llm}=0.5$, and $w_{mllm}=0.3$.

\subsection{Main Results}
Tables~\ref{tab:opening}, \ref{tab:llmibench}, \ref{tab:isg} and \ref{tab:testset} present the detailed results of our model across four distinct benchmarks. Our evaluation compares LLM-I against a diverse set of baselines categorized into three main types: (i) two-stage or compositional methods such as GPT-4o+DALLE-3, Gemini+FLUX/Stable Diffusion 3 (SD3)~\citep{esser2024scaling}, NExT-GPT~\citep{wu2024next}, SEED-X~\citep{ge2024seed}, and ISG~\citep{chen2025interleaved}; (ii) unified models including Show-o (Autoregressive+Diffusion)~\citep{xie2025show} and Anole (Pure Autoregressive)~\citep{chern2024anole}; and (iii) tool-augmented methods, featuring GPT-5~\citep{gpt5} and GPT-4o with a suite of tools that includes search, diffusion, code, and editing capabilities. Across all four benchmarks, we observe that LLM-I exhibits SOTA performance, consistently and significantly outperforming baseline models.

\begin{figure}
    \centering
    \begin{subfigure}[b]{0.322\textwidth}
        \includegraphics[width=\textwidth]{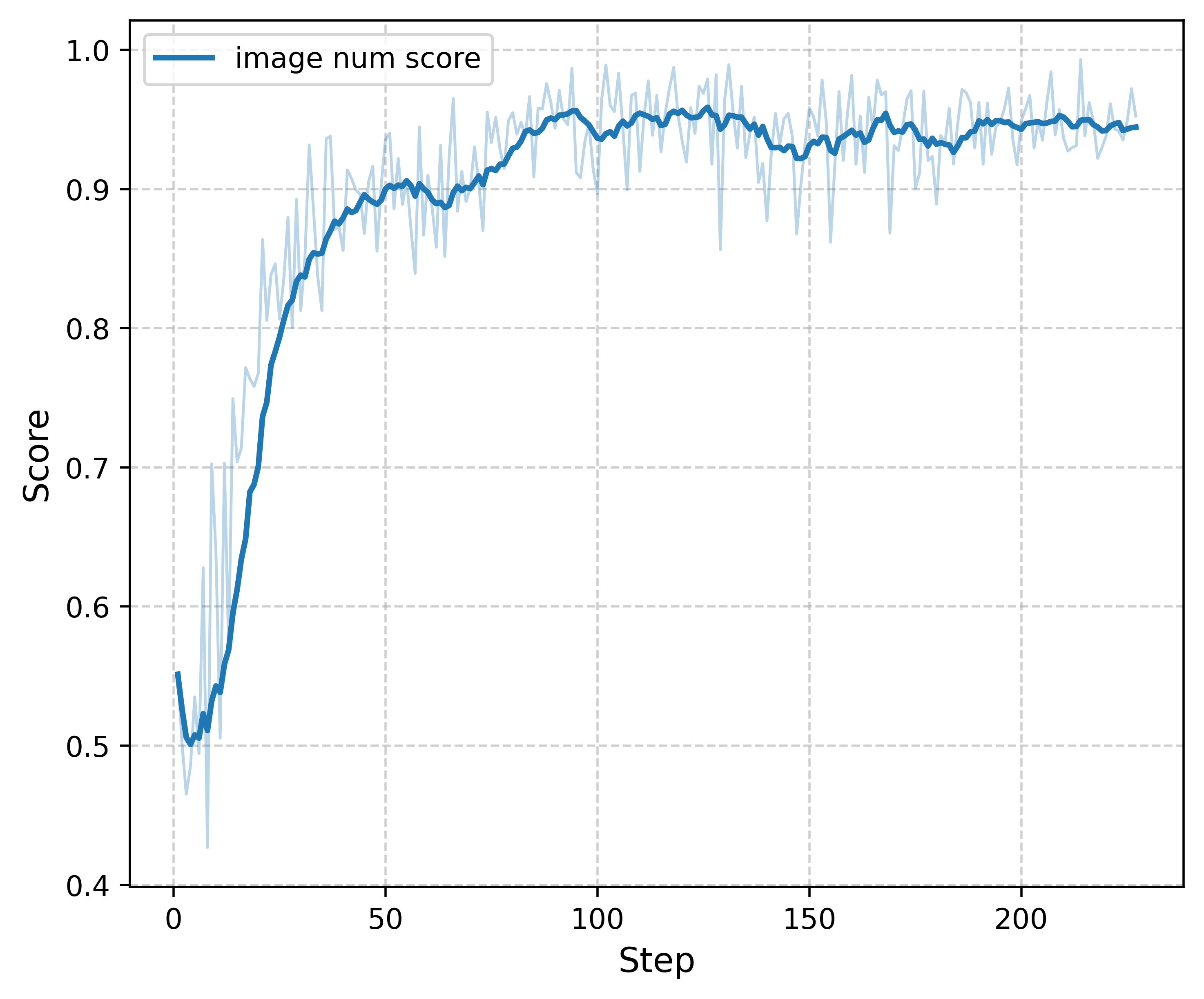}
        \caption{Rule-based Reward}
        \label{fig:metric_curve:a}
    \end{subfigure}
    \hfill 
    \begin{subfigure}[b]{0.322\textwidth}
        \includegraphics[width=\textwidth]{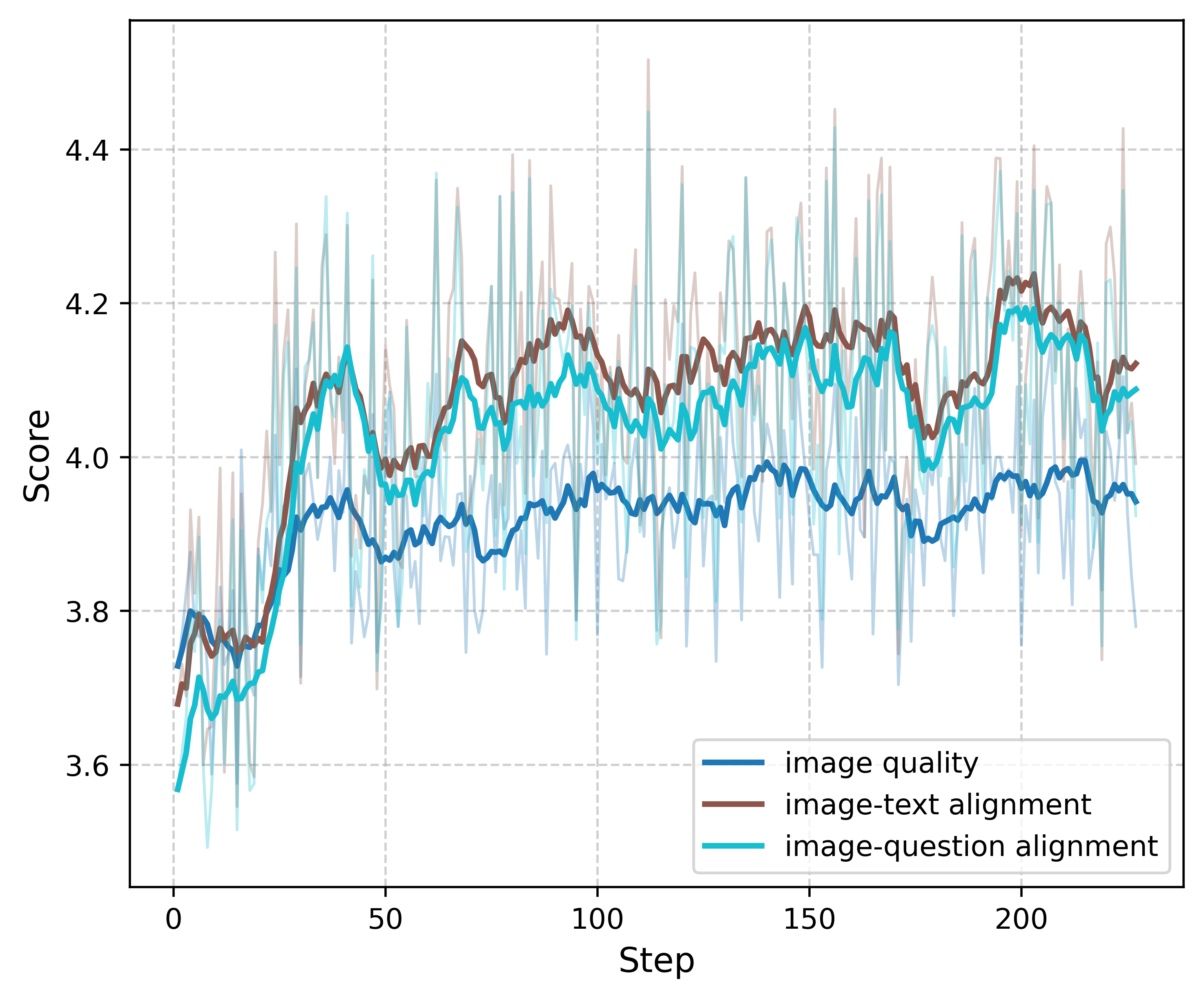}
        \caption{MLLM Judge Reward}
        \label{fig:metric_curve:b}
    \end{subfigure}
    \hfill
    \begin{subfigure}[b]{0.322\textwidth}
        \includegraphics[width=\textwidth]{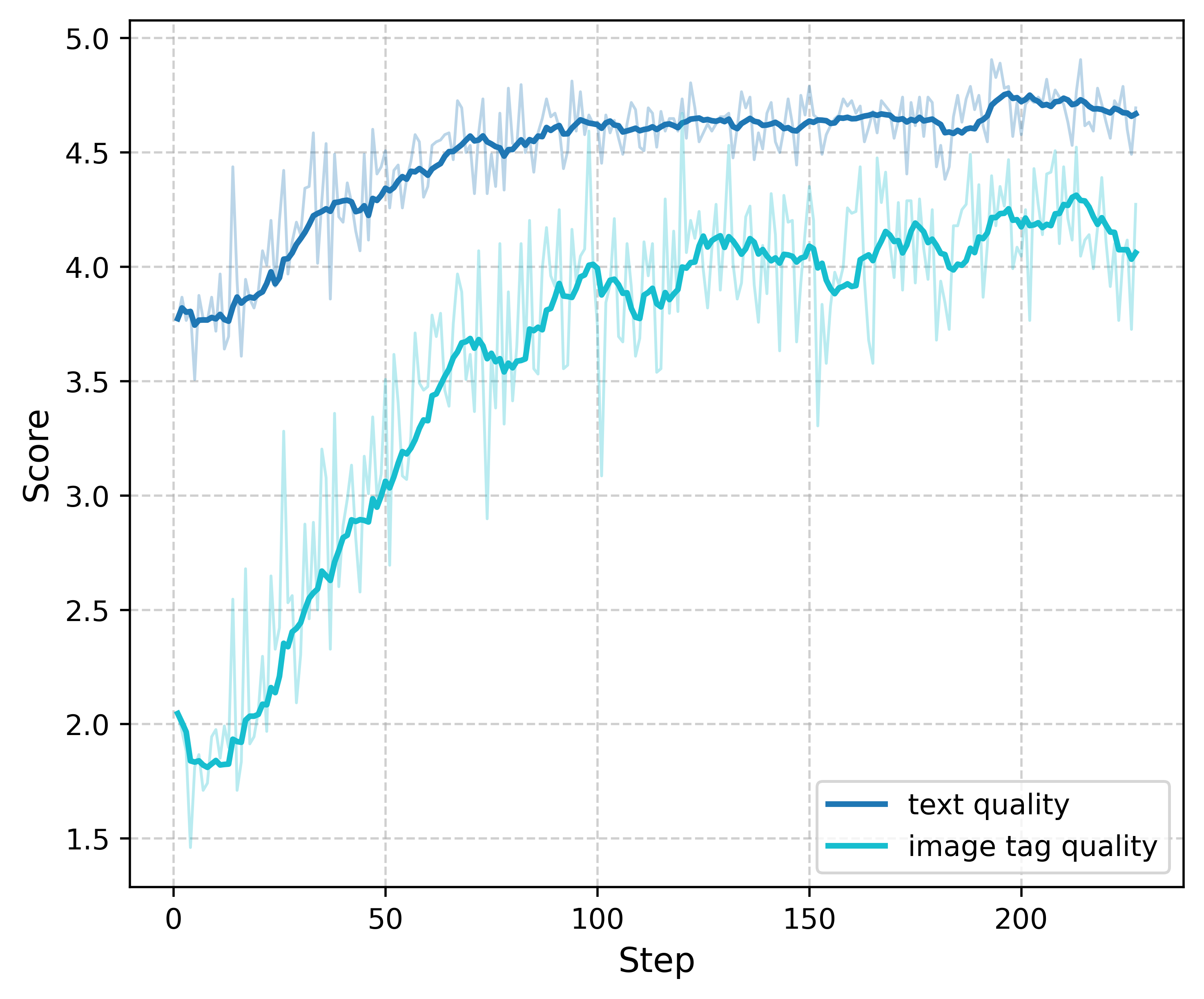}
        \caption{LLM Judge Reward}
        \label{fig:metric_curve:c}
    \end{subfigure}
    \caption{Reward curve during RL training of Qwen2.5-VL-32B.}
    \label{fig:metric_curve}
\end{figure}

\begin{figure}[ht]
    \centering
    \includegraphics[width=\linewidth]{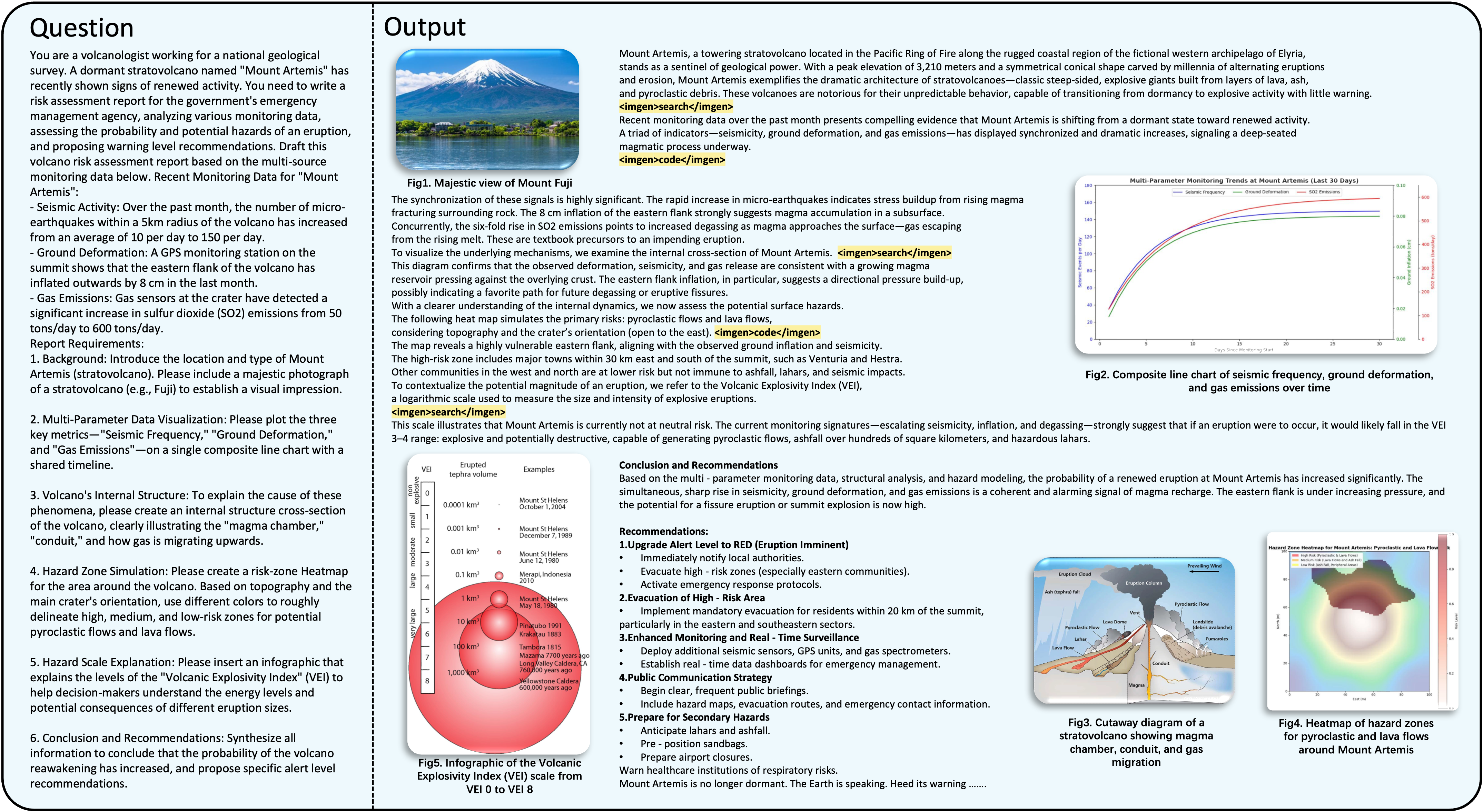}
    \caption{Example generated by LLM-I on LLMI-Bench. Some text is omitted due to space constraints.}
    \label{fig:llmexp1}
\end{figure}

\begin{figure}[ht]
    \centering
    \includegraphics[width=\linewidth]{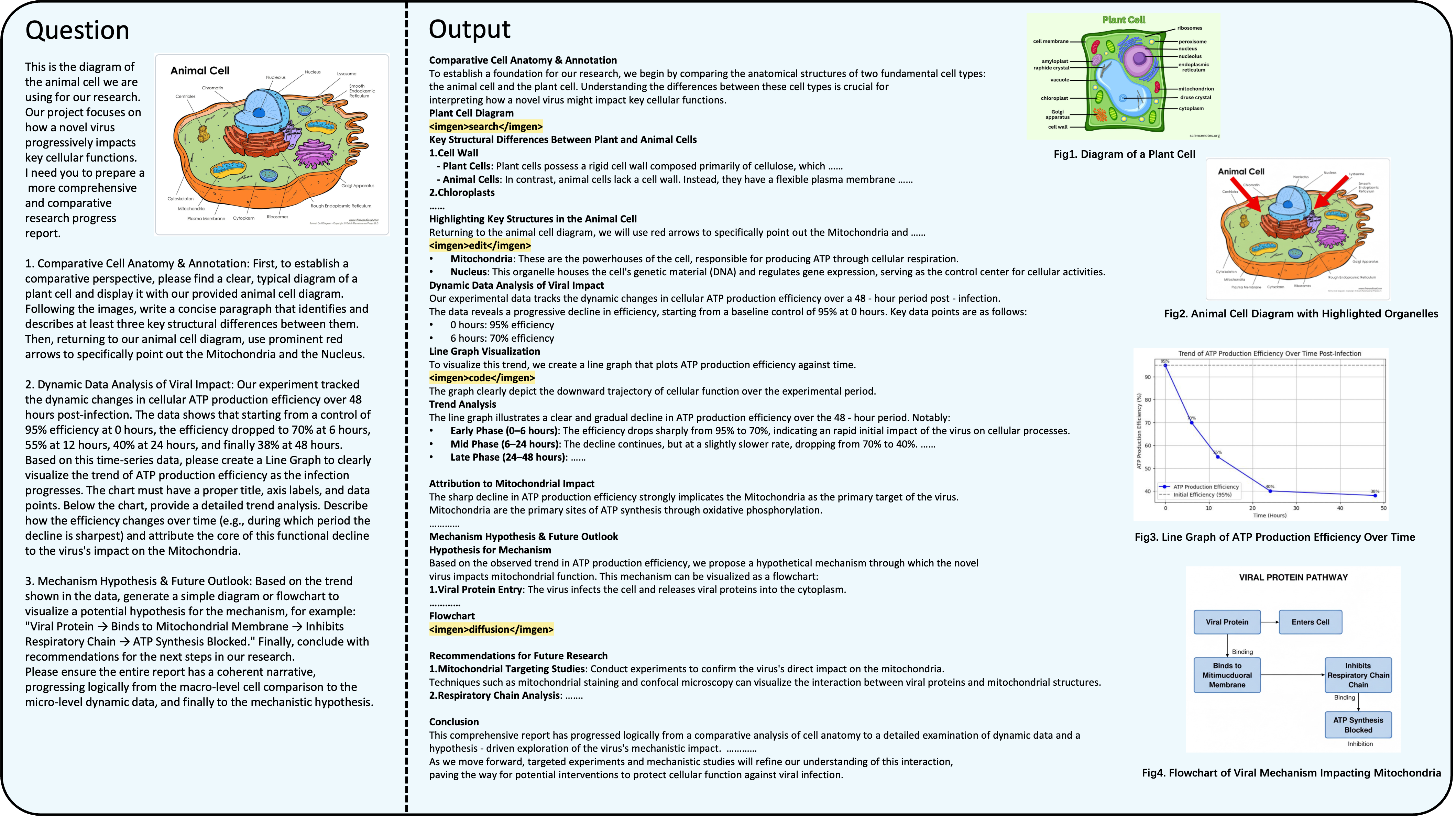}
    \caption{Example generated by MLLM-I on LLMI-Bench. Some text is omitted due to space constraints.}

    \label{fig:mllmexp1}
\end{figure}

\begin{wrapfigure}{r}{5.5cm}
    \centering
    \includegraphics[width=0.33\textwidth]{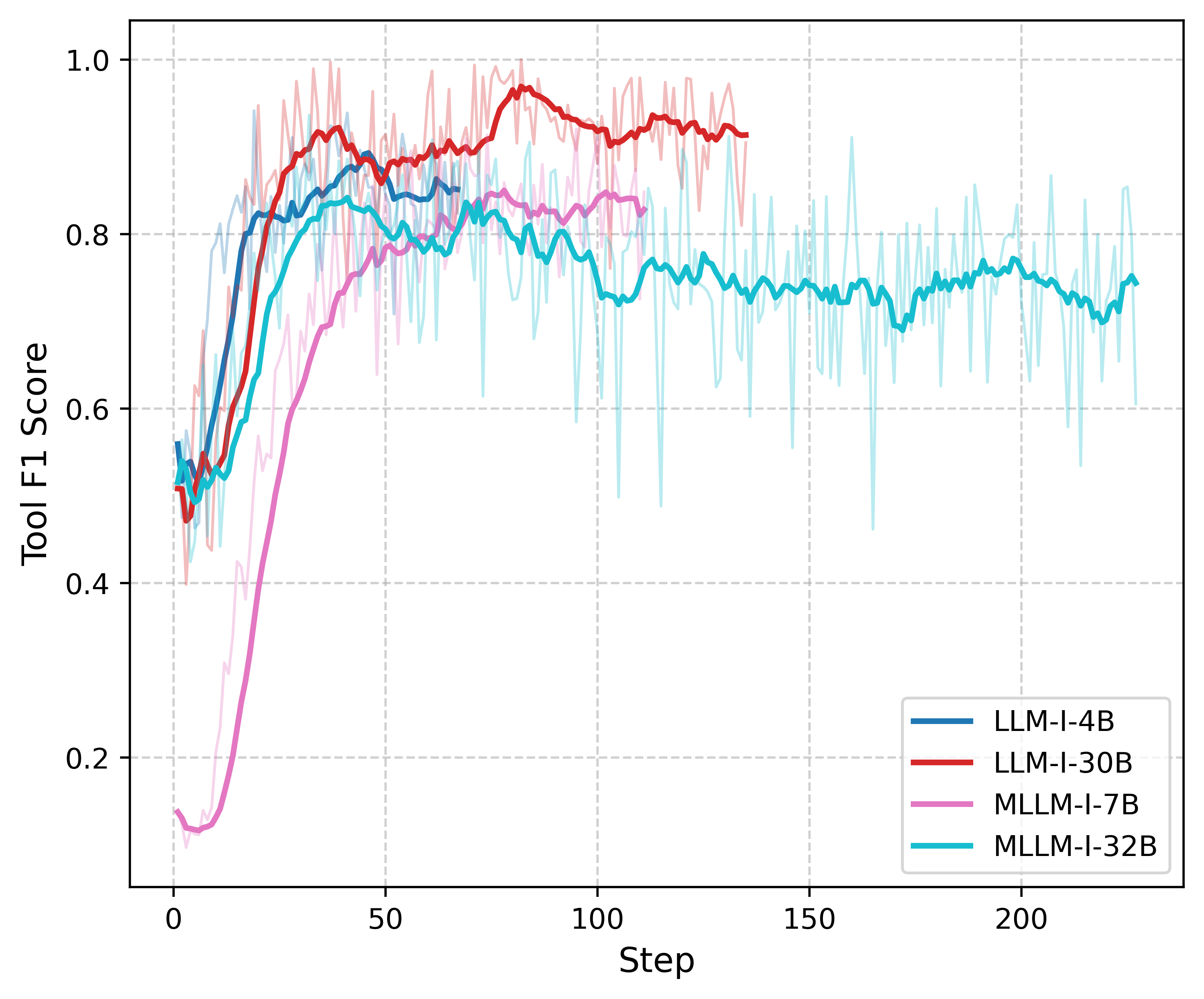}
    \caption{Tool F1 score curve during RL training.}
    \label{fig:toolf1}
    \vskip -0.1in
\end{wrapfigure}

In the general qualitative assessment shown in Table~\ref{tab:opening} and \ref{tab:isg}, the entire LLM-I family shows highly competitive performance, outperforming other leading models and unified approaches, which underscores our model's robustness in generating complete, high-quality, and well-aligned content. This superiority is even more pronounced on specialized benchmarks. On the LLMI-Bench evaluation in Table~\ref{tab:llmibench}, LLM-I models drastically outperform dedicated tool-using agents, including GPT-4o w/Tool and the anticipated GPT-5 w/Tool. This success is largely attributable to our model's exceptional tool invocation capabilities, with LLM-I-4B and LLM-I-30B achieving a perfect 100.0 Tool Accuracy. Additionally, we present the metrics during the RL training in Figure~\ref{fig:metric_curve}. We can observe that with the RL training, the instruction following ability, writing ability and the ability to find images of the model all increases, indicating the effectiveness of RL training.

Furthermore, we present two examples in Figure~\ref{fig:llmexp1} and \ref{fig:mllmexp1}. From the examples, we can observe that LLM-I can intelligently invoke different kinds of tools for image presentation. This shows great advantages over previous methods when requiring real and precise images. In the dataset construction stage in Section~\ref{sec:data}, we define a target tool list for each data item which is verfied by Gemini2.5 Pro and GPT-4o. To further validate the ``intelligence'' of tool invocation, we visualize the tool F1 score in Figure~\ref{fig:toolf1} which evaluates the precision and recall of different tools during the training process. From the figure, the F1 score steadily improves during the RL process, indicating that the model becomes increasingly adept at selecting appropriate tools according to the given context. Notably, no explicit reward is assigned to tool usage; the improvement arises naturally during RL training. This finding suggests that RL not only encourages tool invocation but also enhances the model’s ability to make smarter tool choices for achieving better image–text alignment.

\subsection{Test-Time Scaling}
Table~\ref{tab:tts} presents the performance of our proposed test-time scaling strategy. As detailed in Section~\ref{sec:tts}, this strategy comprises four stages: initial top-k selection, tool enhancement, polishing, and final selection. In our experiments, we set the initial selection parameter $k$ as 4 and sample 2 images for both the search and diffusion tools. The Qwen2.5-VL-72B model serves as both the selector and the polisher.

\begin{wraptable}{r}{0.28\textwidth}
    \centering
    \caption{Results of test-time scaling on LLMI-Bench.}
    \resizebox{0.22\textwidth}{!}{%
        \begin{tabular}{@{}l|cc@{}}
        \toprule
        Model & Rubric & $\Delta$Time \\ \midrule
        LLMI-4B & 88.9 & 0 \\\midrule
        - w stage1 & 91.2 & \textless{}1s \\
        - w stage2 & 91.4 & $\sim$1s \\
        - w stage3 & 89.4 & $\sim$16s/it \\
        w full TTS & 95.1 & $\sim$20s/it \\\midrule
        LLM-I-30B & 94.8 & 0 \\ \bottomrule
        \end{tabular}%
    }
    \label{tab:tts}
\end{wraptable}

The results in Table~\ref{tab:tts} demonstrate the efficacy of our approach. The initial top-k selection and tool enhancement stages substantially boost performance. The subsequent polishing stage also provides improvement. By integrating all four stages, our model surpasses the performance of its 30B counterpart, validating the effectiveness of our test-time scaling strategy.

We also analyze the computational overhead introduced by this strategy. A key advantage is that tool invocations can be processed in parallel. Consequently, the primary overhead consists of only four additional forward passes from the selector/polisher model. The selection process is particularly efficient, as the model only needs to output the optimal index rather than generating a full response. In contrast, the polishing stage is the most time-consuming, as it requires rewriting the entire response.

\begin{table}[]
\centering
\caption{Results of reward ablation experiments on the OpenING benchmark.}
\label{tab:abreward}
\resizebox{\textwidth}{!}{%
\begin{tabular}{@{}l|ccccccc|c@{}}
\toprule
Model & Completeness & Quality & Richness & Correctness & Human Align. & IT Conhe. & MS Consis. & Overall \\ \midrule
Qwen3-4B-Instruct & 6.26 & 6.88 & 5.55 & 6.09 & 6.95 & 5.11 & 6.86 & 6.24 \\\midrule
LLM-I-4B & 8.63 & 8.03 & 7.54 & 8.03 & 8.69 & 7.87 & 8.45 & 8.18 \\\midrule
w/o rule-based reward & 4.22 & 6.55 & 3.93 & 4.55 & 5.68 & 2.34 & 6.05 & 4.76 \\
w/o LLM judge & 8.29 & 7.77 & 7.23 & 7.69 & 8.38 & 7.44 & 8.18 & 7.85 \\
w/o MLLM judge & 8.17 & 7.66 & 7.20 & 7.60 & 8.23 & 7.39 & 8.04 & 7.76 \\ \bottomrule
\end{tabular}%
}
\end{table}

\subsection{Ablation Study}

\textbf{Effectiveness of the Reward Design. }We conduct an ablation study to evaluate the individual contributions of our three reward components: a rule-based reward, an LLM judge, and an MLLM judge. The results are presented in Table~\ref{tab:abreward}. The full LLM-I-4B model, trained on all three rewards, establishes a strong baseline with an overall score of 8.18, demonstrating the effectiveness of the combined reward strategy.

\begin{wrapfigure}{r}{4cm}
    \vskip -0.2in
    \centering
    \includegraphics[width=0.25\textwidth]{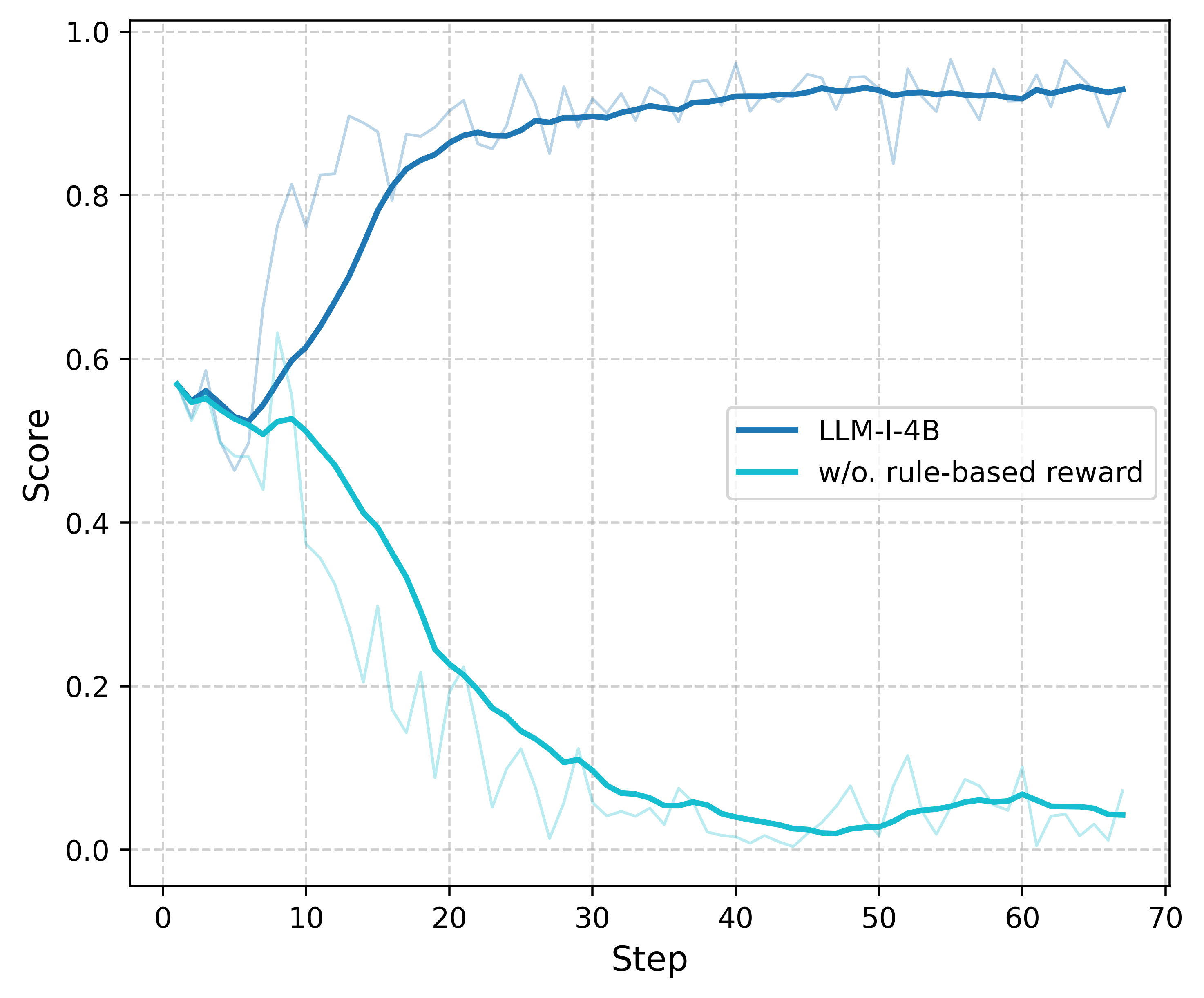}
    \caption{The rule-based reward curve during RL training.}
    \label{fig:abimgnum}
    \vskip -0.1in
\end{wrapfigure}

Particularly, the removal of the rule-based reward proves to be the most detrimental, causing the overall score to plummet to 4.76. As shown in Figure~\ref{fig:abimgnum}, without the rule-based reward, the model will not generate the image to obtain a high score. Comparatively, the performance drop is less severe when removing either the LLM or MLLM judge because their evaluation capabilities likely overlap. Both of these judges assess more nuanced, qualitative aspects of the output like text and image quality. Ultimately, the study confirms that while the rule-based reward provides an essential foundation, the synergistic combination of all three rewards is necessary to achieve the model's peak performance.

\begin{wraptable}{r}{4cm}
    \vskip -0.25in
    \centering
    \caption{Ablation experiments of the tools on LLMI-Bench.}
    \resizebox{0.22\textwidth}{!}{%
    \begin{tabular}{@{}l|c@{}}
    \toprule
    Model & Rubric \\ \midrule
    Qwen3-4B-Instruct wTool & 73.6 \\
    - only diffusion & 66.5 \\
    - only search & 75.2 \\\midrule
    LLM-I-4B & 88.9 \\
    - only diffusion & 76.5 \\
    - only search & 77.5 \\ \bottomrule
    \end{tabular}%
    }
    \label{tab:abtool}
    \vskip -0.2in
\end{wraptable}

\textbf{Tools. }To assess the contribution of individual tools, we perform a tool ablation study and report the results in Table~\ref{tab:abtool}. The results reveal the importance of a comprehensive toolkit, especially for the trained LLMI-4B model. Restricting LLMI-4B to ``only diffusion'' or ``only search'' leads to significant performance degradation. This indicates that its high performance is contingent on its ability to flexibly leverage multiple tools.

Interestingly, the Qwen3-4B model's performance improves to 75.2 when restricted to ``only search'', surpassing its full-toolkit baseline. This counterintuitive result suggests that while the model benefits from the search tool, it may struggle with tool selection when presented with multiple options when it is not trained to use the tools. Forcing it to use only its most effective tool eliminates potential errors in tool orchestration, thereby improving its overall score. From the table, we can observe that the combination of various tools greatly enhances the performance of the model after training. This demonstrates the effectiveness of our ``proficient tool user'' argument for interleaved generation.

\section{Conclusion}
In this paper, we introduce LLM-Interleaved (LLM-I), a framework that overcomes the ``one-tool'' bottleneck in interleaved image-text generation by employing an LLM as an agentic planner. This agent dynamically orchestrates a suite of specialized tools, including web search, diffusion models, code execution, and image editing, to create rich multimodal narratives. LLM-I significantly outperforms state-of-the-art methods, demonstrating that powerful LLMs possess a natural, emergent capability for complex multimodal creation when properly augmented. By championing a flexible "proficient tool-user" paradigm, this work paves the way for future research into expanding the agent's toolkit and enhancing its reasoning for more generalist and capable creative AI.

\clearpage

\bibliographystyle{plainnat}
\bibliography{main}

\clearpage

\beginappendix

\section{Dataset Details}
\label{app:dataset_detail}
In Section~\ref{sec:data}, we detail the design of our dataset, where each item is associated with a required number of images and a corresponding tool list. However, rather than stating these metadata explicitly in the prompt, we guide the model to use specific tools and generate the correct number of images implicitly.

\begin{figure}
    \centering
    \includegraphics[width=0.75\linewidth]{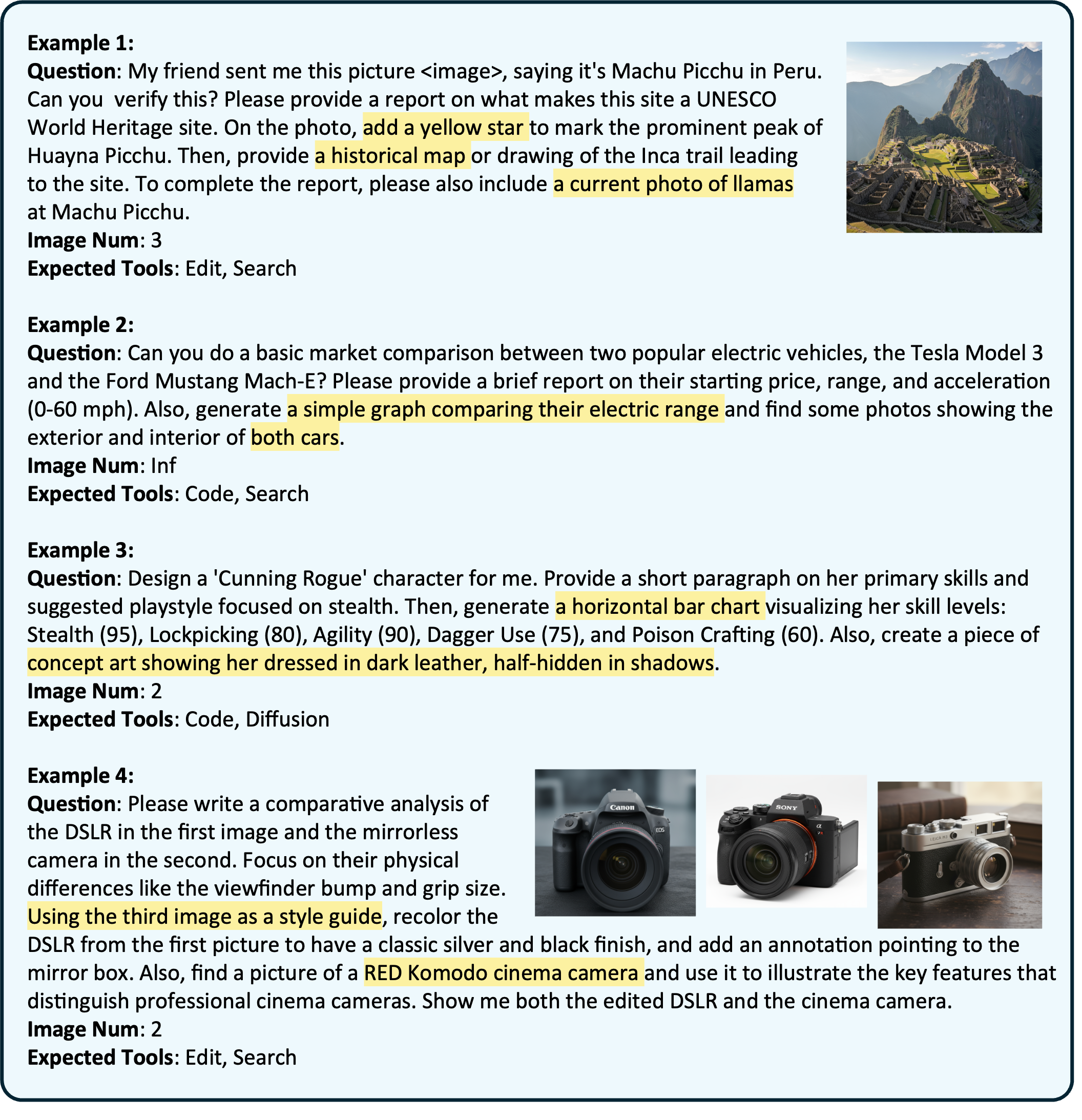}
    \caption{
Examples in our training dataset.
}
    \label{fig:dataexp}
\end{figure}

Figure~\ref{fig:dataexp} presents four examples from our training set, with the text that guides the image generation highlighted in yellow. As shown, the prompts do not explicitly state how to generate the image, but the necessary tools are strongly suggested. For instance, in the first example, the phrase ``add a yellow star to mark...'' implies the need for an image editing tool. Similarly, in the second example, the request for ``a graph comparing the electric range'' suggests using a code interpreter.

\begin{figure}
    \centering
    \includegraphics[width=0.9\linewidth]{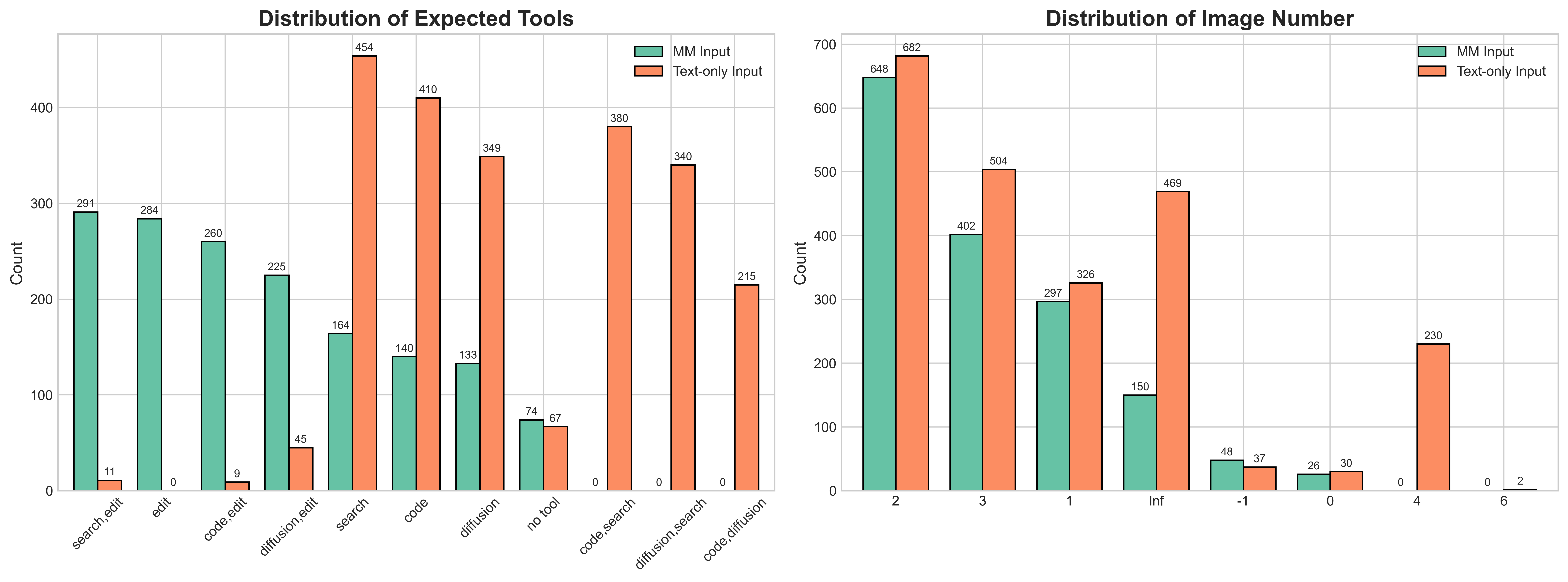}
    \caption{Distribution of our constructed training dataset.}
    \label{fig:datasetdis}
\end{figure}

Furthermore, the required number of images is also not explicitly stated. The model must therefore fully comprehend the prompt's intent to determine the correct number of images to generate. We present the distribution of the datasets in Figure~\ref{fig:datasetdis}.

\begin{figure}[h!]
    \centering
    \includegraphics[width=\linewidth]{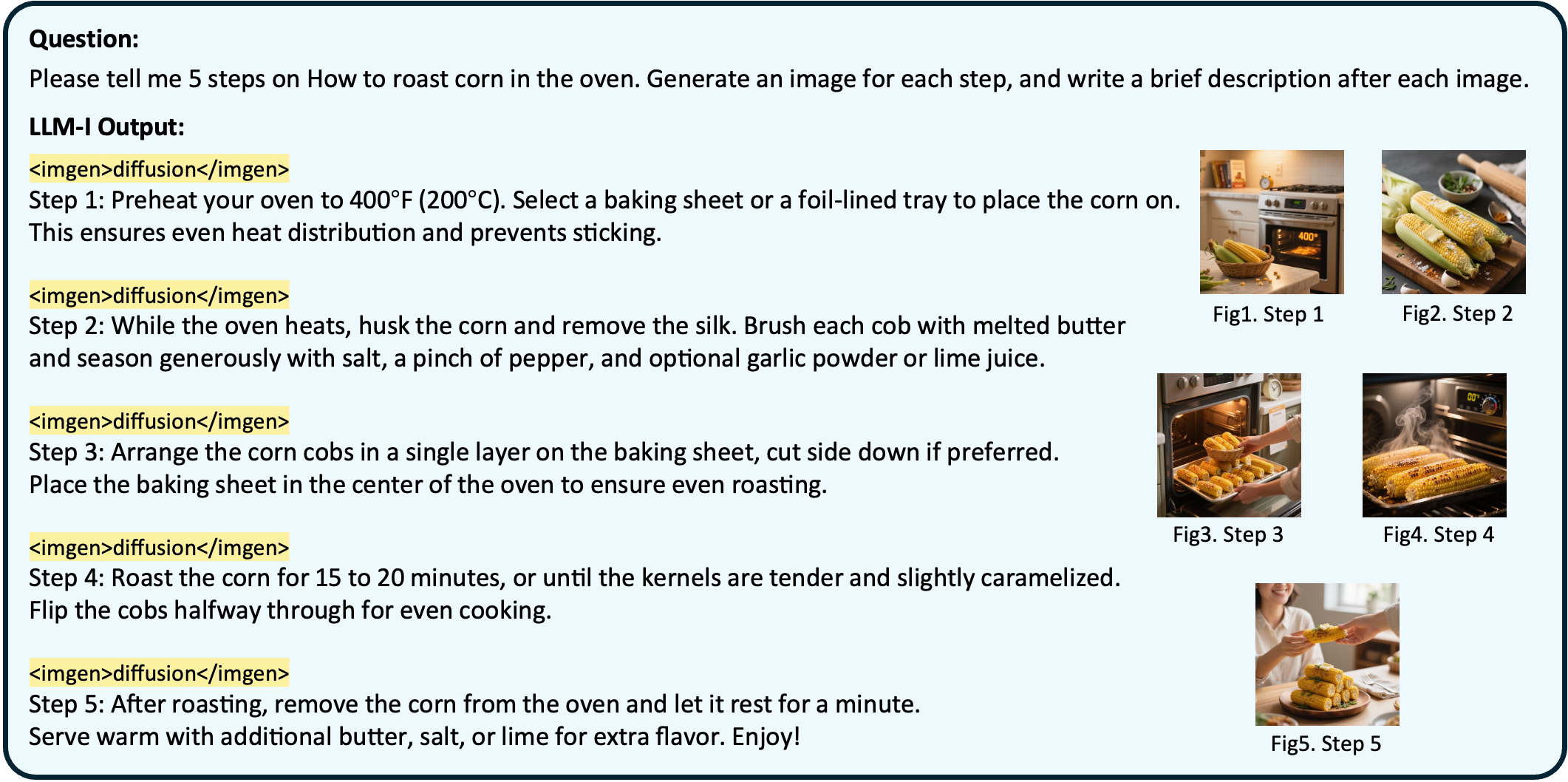}
    \caption{An example generated by LLM-I-4B in the ISG benchmark.}
    \label{fig:isgexp1}
\end{figure}

\section{More Examples}
\begin{figure}[ht]
    \centering
    \includegraphics[width=\linewidth]{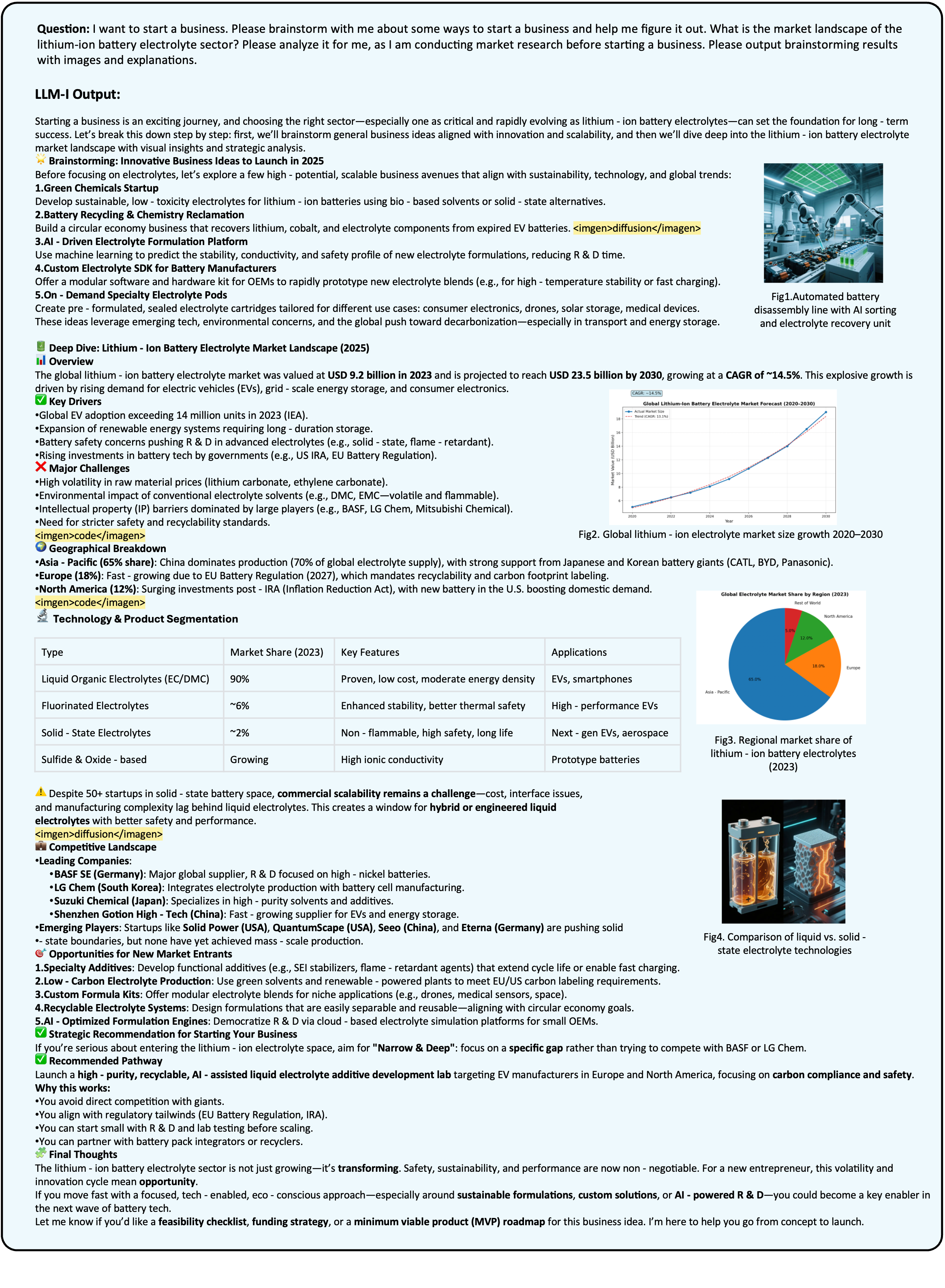}
    \caption{An example generated by LLM-I-30B in the OpenING benchmark.}
    \label{fig:openexp1}
\end{figure}

\begin{figure}[ht]
    \centering
    \includegraphics[width=0.95\linewidth]{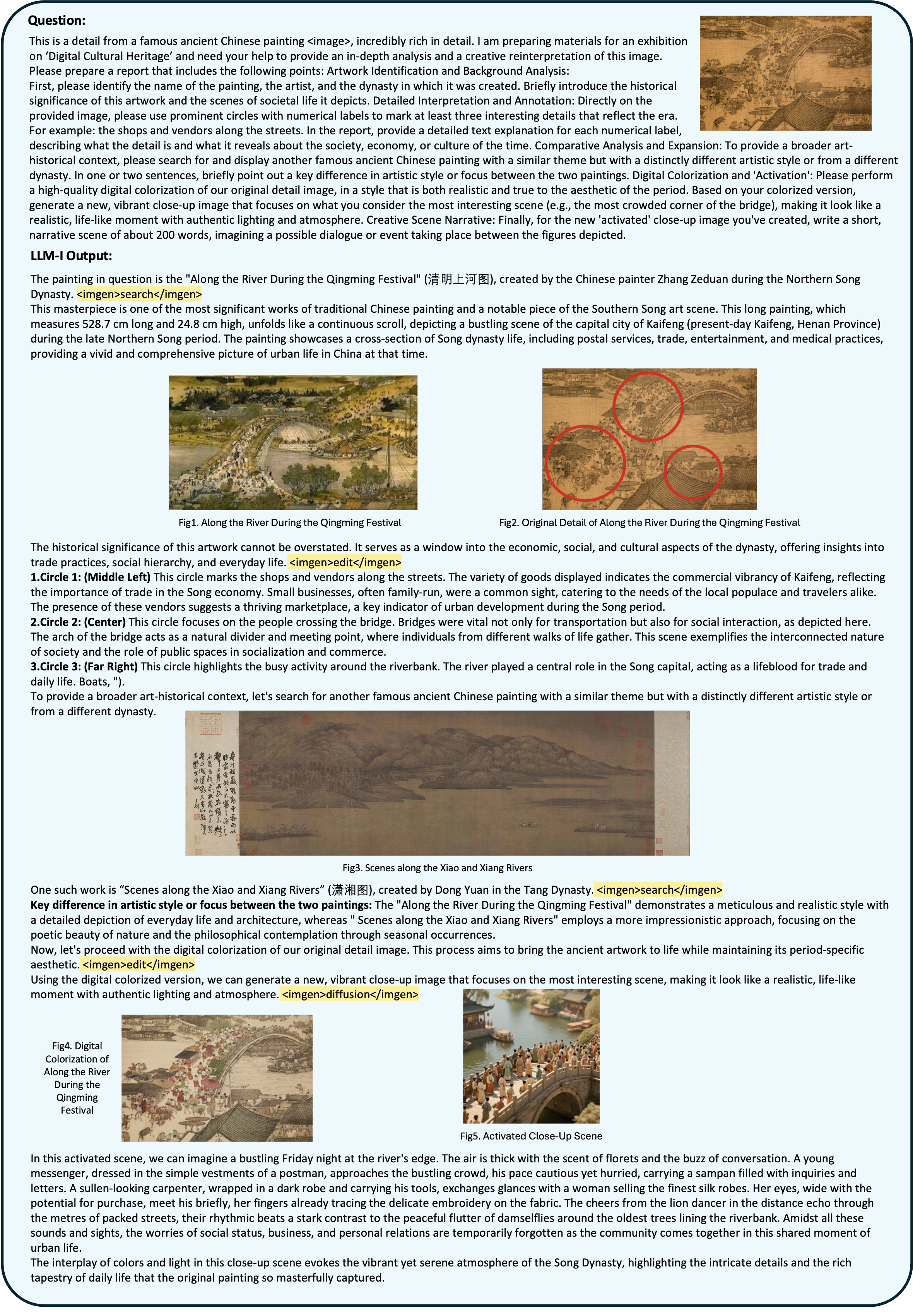}
    \caption{An example generated by MLLM-I-7B in LLMI-Bench.}
    \label{fig:mllmexp2}
\end{figure}

\begin{figure}[ht]
    \centering
    \includegraphics[width=\linewidth]{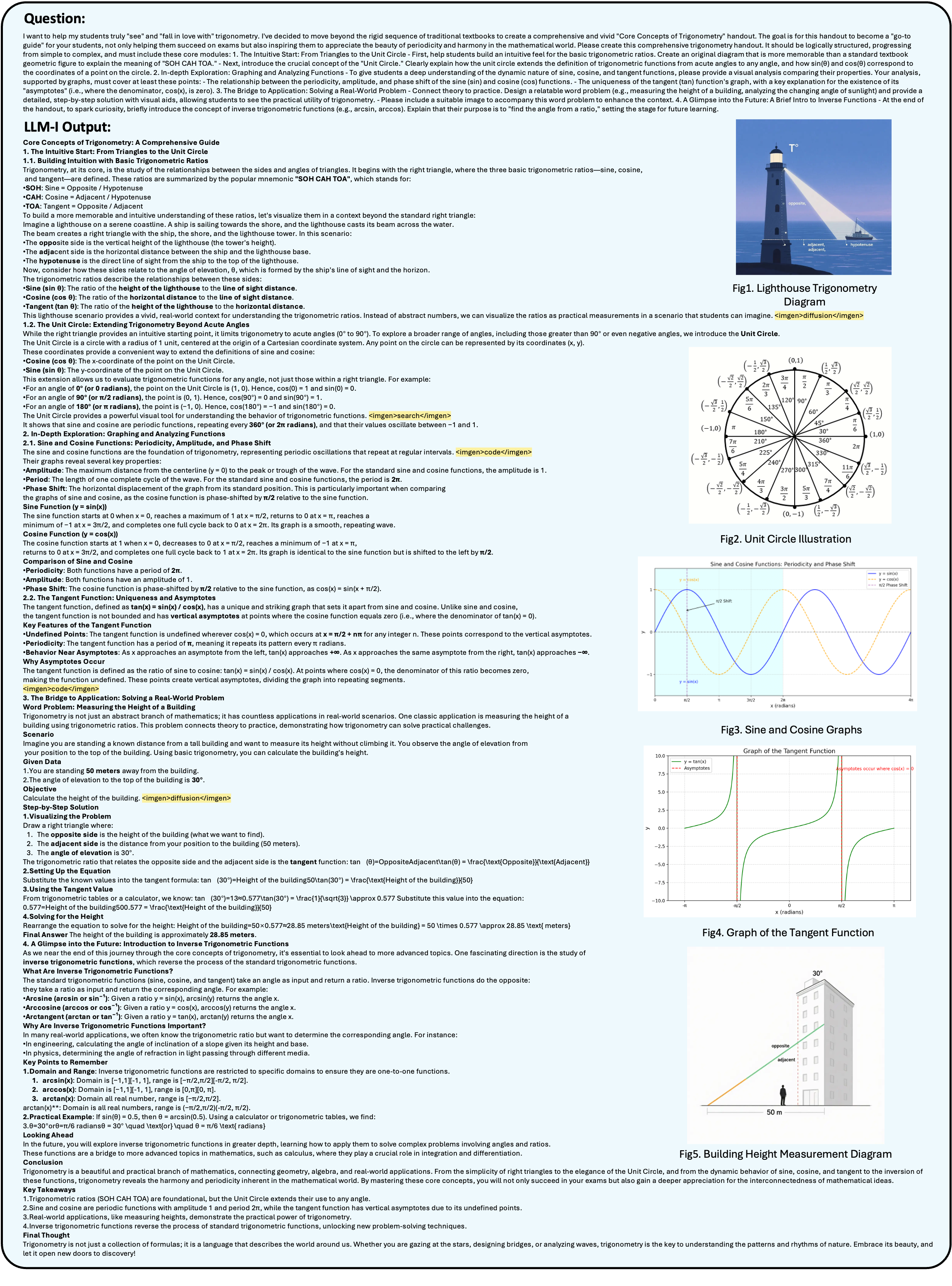}
    \caption{An example generated by MLLM-I-32B in LLMI-Bench.}
    \label{fig:llmexp2}
\end{figure}

To further demonstrate the superiority of LLM-I, we provide additional examples drawn from diverse benchmarks and model backbones. The generated results are shown in Figure~\ref{fig:isgexp1}, \ref{fig:openexp1}, \ref{fig:mllmexp2}, and \ref{fig:llmexp2}. Whether on relatively simple tasks such as the ISG benchmark or OpenING benchmark, or on more challenging tasks such as LLMI-Bench, our method —- scaling from 4B to 32B models —- consistently produces rich, complete responses accompanied by high-quality and highly relevant images. These examples across multiple benchmarks clearly validate both the generalization ability and the superiority of our approach.

\end{document}